\documentclass[conference]{IEEEtran}
\IEEEoverridecommandlockouts
\usepackage{cite}
\usepackage{amsmath,amssymb,amsfonts}
\usepackage{algorithmic}
\usepackage{graphicx}
\usepackage{textcomp}
\usepackage{xcolor}
\usepackage{makecell}
\usepackage{subcaption}
\usepackage{adjustbox}
\usepackage{hyperref}
\usepackage{booktabs}  
\usepackage{multirow}  
\def\BibTeX{{\rm B\kern-.05em{\sc i\kern-.025em b}\kern-.08em
    T\kern-.1667em\lower.7ex\hbox{E}\kern-.125emX}}
\begin{document}

\title{YOLOF-MLD: A Multi-Level Distillation Framework for Efficient Object Detection \\
\thanks{}
}


\author{
    \IEEEauthorblockN{
        Xiaolong Wu$^1$, 
        Liming Zheng$^1$, 
        Zitong Liu$^2$,
        Yue Zheng$^1$
    }
    \IEEEauthorblockA{
        $^1$
        The Chinese University of Hong Kong (Shenzhen), Shenzhen, Guangdong, 518172, P.R. China\\
        $^2$The Chinese University of Hong Kong,
        Shatin, New Territories,
        Hong Kong SAR, China\\
        Email: 224010119@link.cuhk.edu.cn, 125090999@link.cuhk.edu.cn, 1155248030@link.cuhk.edu.hk,
        zhengyue@cuhk.edu.cn
    }
}

\maketitle

\begin{abstract}
Efficient object detection on resource-constrained edge devices is essential for safety-critical applications such as fire-smoke monitoring. However, current lightweight frameworks, such as YOLO26, still rely on the parameter-dense multi-branch Feature Pyramid Network (FPN), preventing full exploitation of their lightweight advantage. Replacing FPN with YOLOF's single-input-single-output (SiSo) neck is a promising solution but causes substantial accuracy loss. To alleviate this, we propose YOLOF-MLD, a CPU-efficient multi-level distillation (MLD) framework that combines a YOLO26n backbone, a SiSo neck, and MLD. It aligns features at the backbone via sliced-semantics matching, neck with receptive-field matching, and response levels. On D-Fire and Annotated Fire-Smoke (AFS) datasets, YOLOF-MLD reduces the parameter count of YOLO26n by 35.7\% and achieves a 62.3\% speedup on Raspberry Pi 5, attaining 71.6\% and 59.9\% mAP\@50, respectively, compared with 75.5\% and 60.0\% for YOLO26n. These results demonstrate a competitive accuracy--latency trade-off for edge deployment. Our code is available at \url{https://github.com/XiaolongWU111/YOLOF---MLD/blob/main/README.md}.

\end{abstract}

\begin{IEEEkeywords}
Lightweight object detection, edge deployment, knowledge distillation, fire-smoke detection
\end{IEEEkeywords}

\section{Introduction}
\label{sec:introduction}
Edge-based visual detection is increasingly required in safety-critical applications such as fire-smoke monitoring and traffic surveillance. Deploying detectors on edge platforms is often constrained by limited computational capability, memory bandwidth, and power budgets. Despite these constraints, models must maintain high accuracy and timely alarms, as missed detections or delayed responses may endanger human lives and cause substantial economic losses. Therefore, developing lightweight, high-performance, and low-latency object detectors has become a critical requirement.

To address this, lightweight detectors have been proposed, which mainly follow two strategies. The first is efficient architecture design (e.g., MobileNet-SSDv2~\cite{1}, NanoDet-Plus~\cite{2}, PP-PicoDet~\cite{3}, and tiny YOLO variants~\cite{4}), which reduces computation but sacrifices representation capability. The second is knowledge distillation (e.g., GID~\cite{5}, SKD~\cite{6}, CrossKD~\cite{7}, and CanKD~\cite{8}), which recovers accuracy without architectural changes, but struggles with large teacher-student feature discrepancies.

To the best of our knowledge, YOLO26\cite{9} achieves the best accuracy–speed trade-off through its optimized architecture design. Nevertheless, YOLO26~\cite{9} still relies on a multi-branch feature pyramid network (FPN)~\cite{13} neck, which utilizes up-sampling, concatenation, and extra convolutions to fuse multi-level features, complicating the topological structure and increasing inference latency.

Replacing FPN~\cite{13} with the YOLOF's\cite{10} single-input-single-output (SiSo) neck can significantly reduce redundant computation and boost inference speed. However, directly coupling the SiSo\cite{10} neck with YOLO26's lightweight backbone and framework is non-trivial. Our empirical observations show that this direct replacement causes a drastic accuracy drop, as the backbone is too lightweight to supply sufficient feature representation, violating YOLOF's ``one-level feature carries enough context" assumption. Furthermore, the conventional YOLOF neck relies on naive dilated convolutions, which lacks the flexibility to adapt to targets with varying scales and irregular shapes, hindering effective context aggregation at the neck level.

To bridge this gap, we propose YOLOF-MLD, a multi-level distillation (MLD) framework tailored for SiSo neck based lightweight detector. By progressively aligning features across the backbone, neck and response levels, our method effectively transfers the rich, multi-scale contextual knowledge of a complex FPN teacher to a highly efficient SiSo student. Our contributions are threefold: 

\begin{enumerate}
\item We propose YOLOF-MLD, a lightweight object detector that replaces the computational expensive multi-branch FPN~\cite{13} neck with an optimized SiSo neck, significantly reducing inference latency on edge devices.
\item We design an MLD framework comprising sliced semantics matching (SSM), receptive field matching (RFM), and response knowledge distillation (RKD) to effectively compensate for the representational loss caused by the architectural simplification.
\item We extensively evaluate the approach on two challenging fire-smoke benchmarks: D-Fire\cite{14} and the annotated fire-smoke (AFS) dataset\cite{15} on CPU platforms. Compared to the YOLO26n~\cite{9} baseline, 
our YOLOF-MLD reduces the parameter count by 35.7\% and achieves relative speedups of 62.3\% and 76.2\% on Raspberry Pi 5 and a 4-thread Intel CPU, respectively, while preserving nearly identical accuracy on AFS (59.9\% vs. 60.0\% mAP50). Despite using more parameters, it also achieves more than 20\% relative speedups than YOLOX-nano~\cite{11} and NanoDet-plus-m~\cite{2} on both platforms. 

\end{enumerate}

\section{Preliminary}
\subsection{FPN Neck \cite{13}}
Multi-scale feature fusion is essential for detecting objects of varying sizes. The mainstream FPN utilizes a multi-branch topology, where high-level semantic features are progressively up-sampled and laterally concatenated with low-level features. Although effective, this architecture introduces complex memory access patterns and significant computational overhead. On resource-constrained edge CPUs, this multi-branch topology severely bottlenecks inference latency.

\subsection{YOLOF's Dilated SiSo Neck \cite{10}}
To eliminate the FPN bottleneck, YOLOF proposes a highly streamlined SiSo neck. Instead of fusing multiple scales, the SiSo neck takes only the deepest feature map from the backbone and processes it through sequentially stacked dilated convolution blocks. By introducing gaps between convolution weights, the SiSo neck expands the receptive field to capture global context without the latency penalty of up-sampling and concatenation.

\section{Methodology}
\label{sec:method}
\begin{figure}[h]
  \centering
  \includegraphics[width=0.49\textwidth]{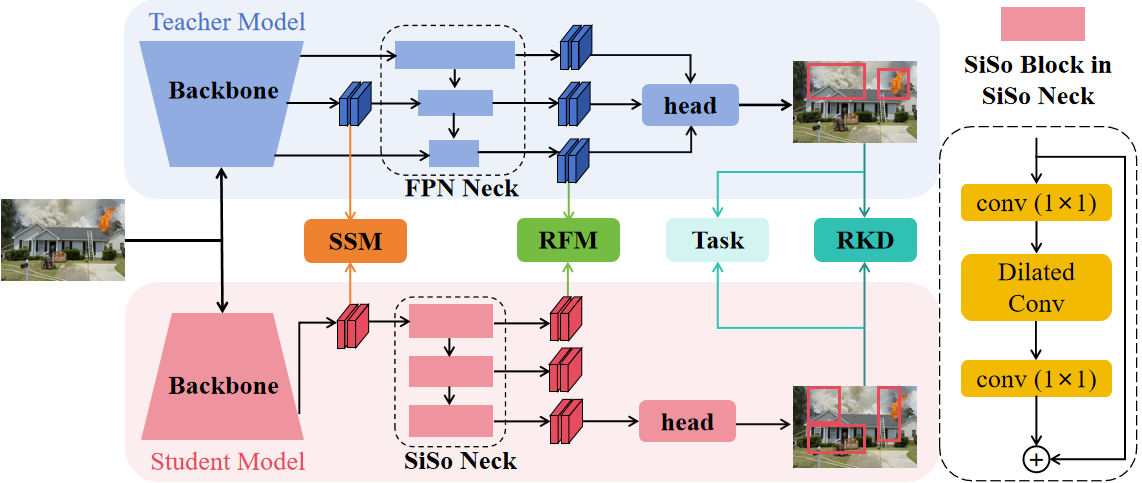}
  \caption{Framework of YOLOF-MLD.}
  \label{fig:framework_od}
\end{figure}

This section details the proposed MLD framework, as illustrated in Fig. \ref{fig:framework_od}. The teacher consists of a backbone $B^T(\cdot)$ and employs a multi-branch FPN~\cite{13} neck $N^T(\cdot)$ that generates three outputs with varying receptive fields. In contrast, the student comprises a backbone $B^S(\cdot)$ and utilizes a lightweight, single-branch SiSo neck $N^S(\cdot)$ comprising three concatenated dilated convolutions. To bridge the resulting representational gap, we propose to align the networks progressively in backbone, neck, and response levels. 

Throughout the paper, superscript $^T$ and $^S$ denote teacher- and student- related variables, respectively, while $c$ and $d$ denote the channel and spatial dimensions of feature maps. 

\vspace{-0.5pt}

    
    

\subsection{Backbone-level Sliced-semantics Matching (SSM)}

Conventional feature distillation typically aligns teacher and student features at corresponding depths. Although teacher's deep layer features provide strong semantic supervision, they have also undergone repeated downsampling and abstraction, and may therefore contain limited fine-grained spatial details. Such details are particular valuable for compensating the representation loss of a compact student. In contrast, the teacher's early-stage features preserve higher-resolution local structures and appearance cues, which motivates us to use an early-stage teacher feature  $f^T \in \mathbb{R}^{c^T \times d^T} $ to supervise the student's late-stage feature $f^S \in \mathbb{R}^{c^S \times d^S}$, enabling the final student representation to retain fine-grained information beyond conventional same-stage distillation.

However, directly aligning $f^T$ and $f^S$ is problematic. The teacher's early-stage feature contains massive, high-dimensional raw information that easily overwhelms the student if forced into a rigid, element-wise alignment. Besides, $f^T$ and $f^S$ differs significantly in both spatial dimensionality and feature distribution. Classical methods\cite{8}\cite{12} attempt to bridge this gap by using parameterized projection layers. 
However, these methods either suffer from the teacher's knowledge loss during the projection process, or induce the student model to rely too much on the projection branch, reducing the gain of distillation. 

To overcome this dimension-distribution mismatch without relying on lossy projections, we propose SSM, which is an optimal transport (OT) inspired distillation. Specifically, given $f^T$ and $f^S$, we first slice them along the channel dimension: $f^T=\left\{f^T_i\right\}_{i=1}^{c^T}$, $f^S=\left\{f^S_j\right\}_{j=1}^{c^S}$. Since the two sliced features differ in spatial dimension, we utilize a lightweight projector $P(\cdot)$ to align the student's spatial resolution to the teacher's, yielding $f_j^{S^\prime}=P(f^S_j) \in \mathbb{R}^{d^T}$, where $P(\cdot)$ is composed only of depth-wise convolutions. This architectural constraint limits the projector's capacity, effectively preventing the student from over-relying on the auxiliary branch. 

Using these aligned features, we establish the matching relationship (i.e., the transport plan) via a joint distribution $\pi(f^S_j, f^T_i)=\pi(f^T_i|f^S_j)\times Pr(f^S_j)$. Since the exact conditional matching probability $\pi(f^T_i|f^S_j)$ is unknown and computationally expensive to strictly solve, we approximate it using a Softmax operation over the cosine similarities between slices. Assuming a uniform marginal distribution for the student slices $Pr(f^S_j)=1/c^S$, the joint distribution is then approximated as:

\vspace{-3pt}
\begin{align} 
\pi(f^S_j, f^T_i) &= \pi(f^T_i|f^S_j) \cdot \frac{1}{c^S}\notag \\
&\approx \frac{\exp(\langle f_j^{S^\prime}, f^T_i \rangle / \tau_1)}{\sum_{k=1}^{c^T} \exp(\langle f_j^{S^\prime}, f^T_k \rangle / \tau_1)}\cdot\frac{1}{c^S} 
\end{align}
where $\tau_1$ is the temperature.
To encourage confident and precise semantic matching, we optimize the projector $P(\cdot)$ via joint entropy minimization:
\begin{align}
J_{pro} &= H(\pi(f^S_j, f^T_i)) \notag \\&= -\Sigma_i\Sigma_j \pi(f^S_j, f^T_i) \cdot \log(\pi(f^S_j, f^T_i))  
\end{align}
By minimizing $J_{pro}$, the projector is forced to discover unambiguous fine-grained co-relationship $\pi(f^S_j, f^T_i)$ between sliced semantics. Consequently, for each student slice $f^S_j$, we can identify its most similar teacher slice $f^T_i$ by maximizing the joint distribution:
\begin{equation}
f^T_i = \mathop{\arg\max}\limits_{f^T_k}\pi(f^S_j, f^T_k)   
\end{equation}

The distillation objective is then formulated to minimize the distance between these optimally matched pairs, which approximates the semi-relaxed Kantorovich OT problem \cite{16}: 
\begin{equation}
J_{SSM} \approx \inf_{\pi \in \Pi} \Sigma_{j=1}^{c^S} \Sigma_{i=1}^{c^T} \|f^{S^\prime}_j - f^T_i\|_2^2 \cdot \pi(f^S_j, f^T_i)
\label{SSM-objective}
\end{equation}
where $\Pi$ denotes the complete set of joint distributions. This OT-inspired formulation provides a highly fine-grained optimization objective without imposing rigid point-to-point constraints. 

However, since the teacher's early-stage features inevitably contain background noise irrelevant to the target task, directly applying Eq. \ref{SSM-objective} is sub-optimal. To alleviate this, we introduce a spatial attention mask $M^T\in\mathbb{R}^{d^T}$ with mean value normalization $V(\cdot)$ to filter out the unrelated noise by averaging the teacher's channels: $M(f^{T}) = \Sigma_i f^T_i/c^T.$
The final spatially-weighted SSM objective becomes:
\begin{equation}
J_{SSM} \approx  \inf_{\pi \in \Pi}\Sigma_j^{c^S} \Sigma_i^{c^T}||f^{S^\prime}_j - f^{T}_i||_2^2 \cdot V(M(f^{T}))\cdot\pi(f^S_j, f^T_i).
\end{equation}

\subsection{Neck-level Receptive-field Matching (RFM)}
Formally, the teacher's FPN neck generates three multi-scale output features $O^T_1, O^T_2, O^T_3$. These features naturally capture contexts with progressively larger receptive fields, i.e., $R(O^T_1) < R(O^T_2) < R(O^T_3)$, where $R(\cdot)$ is the receptive field measure function. In contrast, the student relies on a SiSo neck $N^S(\cdot)$, which sequentially stacks three convolution blocks: $O^S_1=N^S_1(f^S)$, $O^S_2=N^S_2(O^S_1)$ and $O^S_3=N^S_3(O^S_2)$. 

The baseline SiSo design (e.g., YOLOF) relies on standard dilated convolutions. By introducing gaps between convolution weights, dilated convolutions successfully expand the receptive field, achieving $R(O^S_1) < R(O^S_2) < R(O^S_3)$, without adding computational overhead. However, their sampling grids remain strictly rigid and uniformly distributed, hindering the network's ability to model targets with extreme scale variations and irregular geometric deformations. 

To alleviate this, we propose RFM distillation to guide the student's stacked dilated blocks to mimic the FPN's aggregation mechanism. We accomplish this in a divide-and-conquer manner by strictly aligning features with comparable receptive-field scales:
\begin{equation}
J_{RFM} = \Sigma_{i=1}^3||O^T_i - P_i(O^S_i)||_2^2  
\end{equation}
where $P_i(\cdot)$ denotes projectors to match the feature dimensions. By minimizing $J_{RFM}$, the adaptive SiSo neck implicitly acquires the rich multi-scale fusion capability of the FPN, while circumventing the inference latency of a complex multi-branch topology. 

\subsection{Response-level Knowledge Distillation (RKD)}
Beyond intermediate feature alignment, we introduce RKD to directly utilize the teacher's predictions to supervise the student.
Specifically, let $\{I^T_i=(b^T_i, z^T_i)|_{i=1}^{n^T}\}$ and $\{I^S_j=(b^S_j, z^S_j)|_{j=1}^{n^S}\}$ denote the final predictions of the teacher and the student, respectively, where $b$ represents the bounding box coordinates, $z$ is the classification logits, and $n$ is the number of predicted boxes. 

For every student's bounding box, we first find the corresponding teacher's bounding box via the assignment rule $A(\cdot)$ of YOLO26~\cite{9}, i.e, $\{I^T_j=(b^T_{A(j)}, z^T_{A(j)})|_{j=1}^{n^S}\}$. Then we align the two matched predictions. In detail, for bounding box alignment, we utilize IoU loss, and for probability distribution alignment, we utilize KL divergence with temperature $\tau_2$, i.e.,
\begin{align} 
 J_\text{RKD} = \Sigma_{j=1}^{n^S} \Big[ &1 - IoU(b^T_{A(j)}, b^S_j) \notag\\
 &+\tau_2^2 \cdot \text{KL}(z^S_j / \tau_2 \parallel \ z^T_{A(j)} / \tau_2) \Big], 
\end{align} 

\subsection{Online Distillation and Overall Objective}
Finally, we train the entire MLD framework using an online distillation paradigm, optimizing both models in parallel. The total loss is:
\begin{align}
J_{total} = &J^S_{task} + J^T_{task} + (\lambda_{pro}J_{pro} + \lambda_{SSM}J_{SSM} \notag \\ 
&+\lambda_{RFM}J_{RFM} + \lambda_{RKD}J_{RKD}) \cdot s  
\end{align}
where $J^S_{task}$ and $J^T_{task}$ denote the standard detection losses for student and teacher using ground-truth hard labels. The hyperparameters $\lambda_{pro}$, $\lambda_{SSM}$, $\lambda_{RFM}$ and $\lambda_{RKD}$ are empirically set to 0.01, 0.01, 0.008, 0.12 by default to balance the MLD terms, $s$ denotes batch size. 

\section{Experiments}
\subsection{Experiment Settings}
\textbf{Datasets.}
We evaluated our method on 2 datasets: D-Fire\cite{14}, a large-scale dataset comprising 21,527 images of fire and smoke, and Annotated fire-smoke dataset (AFS)~\cite{15}, which is composed of 11027 images of fire and smoke. Both of them effectively simulate harsh real-world environment demanding efficient edge deployment.

\textbf{Implementation Details.} 
We choose YOLO26n as the backbone of both teacher and student, and aim to adapt the FPN-based YOLO26n baseline for efficient edge deployment. The temperatures are empirically set to $\tau_1=0.07$ for SSM and $\tau_2=3$ for RKD. To evaluate inference efficiency across different CPU-based edge settings, we benchmark the models on a Raspberry Pi 5 (Model B Rev 1.1) and an Intel(R) Core(TM) i7-10750H. The latter represents a higher-performance edge gateway. Inference on both platforms is restricted to four CPU threads to emulate constrained per-stream compute resources. Models are trained in Pytorch~\cite{18}, exported and executed via ONNX~\cite{17} Runtime with a batch size of 1.  

\subsection{Performance and Comparison}
To validate the effectiveness of our proposed framework, we compare it against two groups of baselines: 
\begin{itemize} 
\item \textbf{Lightweight Detectors:} Tiny-YOLOv3~\cite{4}, NanoDet-m-0.5x~\cite{2}, NanoDet-Plus-m~\cite{2}, and YOLOX-Nano~\cite{11}. 
\item \textbf{Distillation Methods:} GID~\cite{5}, SKD~\cite{6}, GKD~\cite{12}, CrossKD~\cite{7}, and CanKD~\cite{8}. \end{itemize}

\subsubsection{Comparison with lightweight detectors}
Baseline lightweight detectors are trained using their default settings and official codes, and we set the maximum input image size as 640 under their different multi-scale training strategy.  For inference evaluation, we use an input resolution of $640 \times 640$ and report the frame-per-second (FPS) on the Raspberry Pi and the Intel CPU platforms, measured from model-forward latency only and excluding pre- and post-processing.
For detection accuracy, we adopt the standard mean average precision (mAP) metrics: mAP@0.5 and mAP@0.5-0.95, which are denoted as mAP50 and mAP50-95 for brevity.

Table \ref{tab:main_comparison} compares YOLOF-MLD against existing lightweight detectors across two datasets. Compared with the original YOLO26n baseline, our method reduce the parameter count from 2.38M to 1.53M and achieves a 62.34\% relative speedup (7.86 to 12.76 FPS) on the Raspberry Pi 5 and 76.2\% relative speedup (33.92 to 59.78 FPS) on Intel CPU. This acceleration introduces an accuracy trade-off on D-Fire (mAP50/mAP50-95: 75.5\%/43.4\% to 71.6\%/38.8\%), while nearly preserving the baseline accuracy on AFS (60.0\%/33.5\% vs. 59.9\%/32.6\%). Compared with the ultra-lightweight NanoDet-m-0.5x, YOLOF-MLD prioritizes detection accuracy over maximum throughput, achieving higher mAP50/mAP50-95 on both datasets. Moreover, despite having more parameters, YOLOF-MLD outperforms YOLOX-nano and NanoDet-plus-m in inference speed, achieving relative speed up of 26.8\%/20.57\% and 32.9\%/46.99\% on Raspberry Pi 5/Intel CPU, respectively. These results suggest that parameter count alone does not characterize practical CPU inference efficiency, and a streamlined SiSo topology can offer favorable balance between detection accuracy and deployment speed.

\begin{table}[t]
\centering
\caption{Comparisons with Lightweight Object Detectors}
\label{tab:main_comparison}
\footnotesize 
\setlength{\tabcolsep}{2.0pt} 
\resizebox{\columnwidth}{!}{%
\begin{tabular}{lccccccc} 
\toprule
\multirow{2}{*}{\textbf{Model}} & \multirow{2}{*}{\textbf{Param.}} & \multicolumn{2}{c}{\textbf{FPS (Avg)}} & \multicolumn{2}{c}{\textbf{D-Fire [14]}} & \multicolumn{2}{c}{\textbf{AFS [16]}} \\
\cmidrule(lr){3-4} \cmidrule(lr){5-6} \cmidrule(lr){7-8}
 & & \textbf{Raspberry Pi} & \textbf{Intel CPU} & \textbf{mAP50} & \textbf{mAP50-95} & \textbf{mAP50} & \textbf{mAP50-95} \\
\midrule
TN-YOLOv3 \cite{4}       & 12.13M & 3.79  & 15.81 & 71.5 & 37.1 & 54.7 & 28.3 \\
NanoDet-m-0.5x \cite{2}  & 0.26M  & 26.23 & 97.67 & 70.8 & 33.5 & 56.2 & 27.2 \\
NanoDet-plus-m \cite{2}  & 1.17M  & 9.60  & 40.67 & 77.8 & 43.9 & 56.4 & 32.1 \\
YOLOX-nano \cite{11}     & 0.90M  & 10.06 & 49.58 & 72.2 & 41.1 & 57.8 & 32.0 \\
YOLO26n \cite{9}         & 2.38M  & 7.86  & 33.92 & 75.5 & 43.4 & 60.0 & 33.5 \\
\textbf{YOLOF-MLD (Ours)} & 1.53M  & 12.76 & 59.78 & 71.6 & 38.8 & 59.9 & 32.6 \\
\bottomrule
\end{tabular}%
} 

\vspace{6pt}
\parbox{\columnwidth}{\scriptsize \textit{Note:} FPS (Avg) is the average FPS across both datasets, excluding pre-processing and post-processing (e.g., non-max suppression). }
\end{table}


\subsubsection{Comparison with distillation methods}
For all distillation-related experiments, we adopt a unified training configuration for fair comparison. Specifically, the baseline and competing KD methods (GID, SKD, GKD, CrossKD, and CanKD) are trained for 200 epochs with a batch size of 32 using the MuSGD optimizer, an initial learning rate of 0.01 (for CanKD\cite{8}, the learning rate is set as 0.001 for numerical stability), and a linear decay schedule. All methods use the same student architecture, consisting of the YOLO26n~\cite{9} backbone and YOLOF\cite{10}'s dilated SiSo neck, and follow the original YOLO26 label-assignment rule.

\begin{table}[t]
\centering
\caption{Comparison of Distillation and Ablation Study}
\label{tab:kd_comparison}
\footnotesize 
\setlength{\tabcolsep}{1.5pt} 
\begin{tabular}{lcccc}
\toprule
\multirow{2}{*}{\textbf{Method}} & \multicolumn{2}{c}{\textbf{D-Fire \cite{14}}} & \multicolumn{2}{c}{\textbf{AFS \cite{15}}} \\
\cmidrule(lr){2-3} \cmidrule(lr){4-5}
 & \textbf{mAP50} & \textbf{mAP50-95} & \textbf{mAP50} & \textbf{mAP50-95} \\
\midrule
Baseline$^1$ &69.8  &37.8  &58.8  &31.6  \\
\midrule
GID \cite{5}          &  70.0{\scriptsize(+0.2)} &37.8  {\scriptsize(+0.0)} &58.2  {\scriptsize(-0.6)} &31.7  {\scriptsize(+0.1)} \\
SKD \cite{6}          & 70.3 {\scriptsize(+0.5)} & 37.3 {\scriptsize(-0.5)} & 60.3 {\scriptsize(+1.5)} &32.4  {\scriptsize(+0.8)} \\
GKD \cite{12}          &69.5  {\scriptsize(-0.3)} & 36.4 {\scriptsize(-1.4)} &55.8  {\scriptsize(-3.0)} &29.5  {\scriptsize(-2.1)} \\
CrossKD \cite{7}      &70.2  {\scriptsize(+0.4)} &38.0  {\scriptsize(+0.2)} &59.2  {\scriptsize(+0.4)} &31.7  {\scriptsize(+0.1)} \\
CanKD \cite{8}        &70.4  {\scriptsize(+0.6)} &38.0 {\scriptsize(+0.2)} &59.5  {\scriptsize(+0.7)} &31.9  {\scriptsize(+0.3)} \\
\midrule
+ Uniform Matching\cite{10}$^2$  &  65.9{\scriptsize(-3.9)} &34.9  {\scriptsize(-2.9)} &  58.6{\scriptsize(-0.2)} &  30.0{\scriptsize(-1.6)} \\
+ RFM             &70.2  {\scriptsize(+0.4)} &38.1  {\scriptsize(+0.3)} &58.4  {\scriptsize(-0.4)} &31.8  {\scriptsize(+0.2)} \\
+ RFM + SSM             &70.5  {\scriptsize(+0.7)} &37.8  {\scriptsize(+0.0)} &59.7  {\scriptsize(+0.9)} &32.2  {\scriptsize(+0.6)} \\
+ RFM + SSM + RKD(ours)             &71.6  {\scriptsize(+1.8)} &38.8  {\scriptsize(+1.0)} &59.9  {\scriptsize(+1.1)} &32.6  {\scriptsize(+1.0)} \\
\bottomrule
\end{tabular}
\parbox{\columnwidth}{\scriptsize \textit{Note:} $^1$Student without distillation; $^2$ Top k=4.}
\end{table}


As shown in Table \ref{tab:kd_comparison}, most general-purpose methods provide weak improvements on the baselines and may even cause negative transfer; for example, SKD improves the baseline by 1.5 percentage points of mAP50 on AFS but decreases by 0.5 percentage points of mAP50-95 on D-Fire. In contrast, our MLD strategy consistently improves the baseline by 1.8/1.0 mAP50/mAP50-95 on D-Fire and 1.1/1.0 on AFS. Notably, it also outperforms the strongest competitor (CanKD) by a significant margin of +0.8 and +0.7 mAP50-95 on D-Fire and AFS, respectively, demonstrating the effectiveness of the proposed multi-level transfer for the FPN-to-SiSo setting.

\subsection{Ablation Study}
We conduct ablations on the proposed YOLOF-MLD framework. The baseline consists of the YOLO26n backbone and the SiSo neck, trained without distillation while retaining the original YOLO26n dynamic label-assignment rule. We first replace this rule with YOLOF's Uniform Matching~\cite{10}, which encourages a more balanced positive-negative label assignment ratio to compensate for the accuracy drop caused by replacing FPN with dilated SiSo neck. However, it reduces overall performance on D-Fire datasets by up to 3.9 mAP50 and on AFS datasets by up to 1.6 mAP50-95, possibly because sparse prediction locations on low-resolution SiSo feature maps may yield low-quality positive samples with substantial background content.
We then add RFM, SSM, and RKD individually to the baseline. As shown in Table~\ref{tab:kd_comparison}, every sub-method continues to improve the baselines; combining all components yields a cumulative gain of 1.8/1.1 mAP50 and 1.1/1.0 mAP50-95. 
 

\subsection{Conclusion and Future Works}
In conclusion, we presented YOLOF-MLD, a CPU-efficient object detector that combines a YOLO26n backbone with a lightweight SiSo neck derived from YOLOF. By incorporating multi-level knowledge distillation, the proposed framework mitigates the accuracy degradation caused by replacing the multi-branch FPN neck. Experiments on two fire-smoke detection datasets demonstrate that YOLOF-MLD substantially improves the inference efficiency of the YOLO26n baseline on both Raspberry Pi and thread-constrained Intel CPU,  while maintaining competitive detection accuracy. In future work, we will further investigate backbone optimization and its integration with the proposed MLD framework to reduce remaining computational redundancy.

\color{red}


\begin{thebibliography}{00}
\bibitem{1} Y.-C. Chiu, C.-Y. Tsai, M.-D. Ruan, G.-Y. Shen, and T.-T. Lee, “Mobilenet-SSDv2: An improved object detection model for embedded systems,” in 2020 International conference on system science and engineering (ICSSE), 2020, pp. 1–5.
\bibitem{2} RangiLyu, “NanoDet-Plus: Super fast and high accuracy lightweight anchor-free object detection model,” 2021. [Online]. Available: https://github.com/RangiLyu/nanodet
\bibitem{3} G. Yu et al., “PP-PicoDet: A better real-time object detector on mobile devices,” arXiv preprint arXiv:2111.00902, 2021.

\bibitem{4} P. Adarsh, P. Rathi, and M. Kumar, “YOLO v3-Tiny: Object Detection and Recognition using one stage improved model,” in 2020 6th international conference on advanced computing and communication systems (ICACCS), 2020, pp. 687–694. 

\bibitem{5}X. Dai et al., “General instance distillation for object detection,” in 2021 IEEE/CVF Conference on Computer Vision and Pattern Recognition (CVPR), IEEE, 2021, pp. 7838–7847.

\bibitem{6}P. De Rijk, L. Schneider, M. Cordts, and D. Gavrila, “Structural knowledge distillation for object detection,” Advances in Neural Information Processing Systems, vol. 35, pp. 3858–3870, 2022.

\bibitem{7}J. Wang, Y. Chen, Z. Zheng, X. Li, M.-M. Cheng, and Q. Hou, “CrossKD: Cross-head knowledge distillation for object detection,” in 2024 IEEE/CVF Conference on Computer Vision and Pattern Recognition (CVPR), IEEE, 2024, pp. 16520–16530.

\bibitem{8}S. Sun and W. Ohyama, “CanKD: Cross-Attention-based Non-local operation for Feature-based Knowledge Distillation,” in 2026 IEEE/CVF Winter Conference on Applications of Computer Vision (WACV), IEEE, 2026, pp. 8606–8616.

\bibitem{9} G. Jocher, J. Qiu, M. Liu, S. Lyu, F. C. Akyon, and M. E. Kalfaoglu, “Ultralytics YOLO26: Unified real-time end-to-end vision models,” arXiv preprint arXiv:2606.03748, 2026.

\bibitem{10} Q. Chen, Y. Wang, T. Yang, X. Zhang, J. Cheng, and J. Sun, “You only look one-level feature,” in 2021 IEEE/CVF conference on computer vision and pattern recognition (CVPR), 2021, pp. 13034–13043.


\bibitem{11}Z. Ge, S. Liu, F. Wang, Z. Li, and J. Sun, “YOLOX: Exceeding YOLO series in 2021,” arXiv preprint arXiv:2107.08430, 2021.


\bibitem{12}Q. Lan and Q. Tian, “Gradient-guided knowledge distillation for object detectors,” in 2024 IEEE/CVF Winter Conference on Applications of Computer Vision (WACV), 2024, pp. 423–432.

\bibitem{13}T.-Y. Lin, P. Dollár, R. Girshick, K. He, B. Hariharan, and S. Belongie, “Feature pyramid networks for object detection,” in 2017 IEEE conference on computer vision and pattern recognition (CVPR), 2017, pp. 936–944.

\bibitem{14}P. V. A. Venancio, A. C. Lisboa, and A. V. Barbosa, “An automatic fire detection system based on deep convolutional neural networks for low-power, resource-constrained devices,” Neural Computing and Applications, vol. 34, no. 18, pp. 15349–15368, 2022.


\bibitem{15}S. Partheepan, “Annotated Fire -Smoke Image Dataset for fire detection Using YOLO,” 4 2025, doi: 10.25946/28747046.v1.

\bibitem{16}C. Villani, “Optimal transport: old and new,” vol. 338, 2009.

\bibitem{17}O. R. developers, “ONNX Runtime.” Accessed: Oct. 22, 2026. [Online]. Available: https://onnxruntime.ai/

\bibitem{18}A. Paszke et al., “Pytorch: An imperative style, high-performance deep learning library,” Advances in neural information processing systems, vol. 32, 2019.

\end{thebibliography}
\end{document}